\documentclass[letterpaper]{article} 
\usepackage{aaai25}  
\usepackage{times}  
\usepackage{helvet}  
\usepackage{courier}  
\usepackage[hyphens]{url}  
\usepackage{graphicx} 
\usepackage{natbib}  
\usepackage{caption} 
\nocopyright
\usepackage{multirow}
\usepackage{graphicx}
\usepackage{amsmath}
\usepackage{float}

\usepackage{array}
\usepackage{booktabs}       
\usepackage{amsfonts}       
\usepackage{nicefrac}       
\usepackage{microtype}      
\usepackage{xcolor}         

\usepackage{algorithm}
\usepackage{algorithmic}

\usepackage{newfloat}
\usepackage{listings}
\DeclareCaptionStyle{ruled}{labelfont=normalfont,labelsep=colon,strut=off} 
\floatstyle{ruled}
\newfloat{listing}{tb}{lst}{}
\floatname{listing}{Listing}
\title{BrainVis: Exploring the Bridge between Brain and Visual Signals via Image Reconstruction}
\author{
    Honghao Fu\textsuperscript{\rm 1, \rm 2}, Zhiqi Shen\textsuperscript{\rm 2}, Jing Jih Chin\textsuperscript{\rm 2}, Hao Wang\textsuperscript{\rm 1}\thanks{Corresponding Author}\\
}
\affiliations{
    \textsuperscript{\rm 1}The Hong Kong University of Science and Technology (Guangzhou), China\\
    \textsuperscript{\rm 2}Nanyang Technological University, Singapore\\

}

\usepackage{bibentry}
\begin{document}
\maketitle

\begin{abstract}
Analyzing and reconstructing visual stimuli from brain signals effectively advances the understanding of human visual system. However, the EEG signals are complex and contain significant noise. This leads to substantial limitations in existing works of visual stimuli reconstruction from EEG, such as difficulties in aligning EEG embeddings with the fine-grained semantic information and a heavy reliance on additional large self-collected dataset for training. To address these challenges, we propose a novel approach called BrainVis. Firstly, we divide the EEG signals into various units and apply a self-supervised approach on them to obtain EEG time-domain features, in an attempt to ease the training difficulty. Additionally, we also propose to utilize the frequency-domain features to enhance the EEG representations. Then, we simultaneously align EEG time-frequency embeddings with the interpolation of the coarse and fine-grained semantics in the CLIP space, to highlight the primary visual components and reduce the cross-modal alignment difficulty. Finally, we adopt the cascaded diffusion models to reconstruct images. Using only 10\% training data of the previous work, our proposed BrainVis outperforms state of the arts in both semantic fidelity reconstruction and generation quality. The code is available at \url{https://github.com/RomGai/BrainVis}.

\end{abstract}

%

\section{Introduction}
\label{sec:intro}

Advancements in deep learning have markedly enhanced the ability to decode complex neural signals~\cite{supply-2}, laying the groundwork for studies investigating neural responses to external stimuli~\cite{208, 101, 102}. This body of research extends to exploring the nexus of neural signals with sensory perceptions such as auditory, olfactory, and visual stimuli~\cite{103, supply-3, 223}. Particularly, investigating how the brain processes visual information is paramount for advancing our comprehension of the visual functions of the human brain. Nevertheless, deciphering the brain's visual system is complicated by the extensive involvement of neurons~\cite{104}.

In response, recent investigations have enlisted deep neural networks for the extraction of visual information from neural signals, reconstructing images semantically related to these signals to explore the visual semantic components contained in neural signals~\cite{207, 208, 210, 211, 218, 220, 221}.  Electroencephalography (EEG), capturing general brain activity via scalp electrodes to generate multi-channel 2D temporal sequences~\cite{223}, stands out for its affordability and simplicity in processing~\cite{108}. Despite these benefits, EEG's limited spatial resolution, the challenge of accurately localizing brain regions, and susceptibility to extraneous noise make reconstructing images from EEG signals a formidable task~\cite{109}.

\begin{figure}[t]
  \centering
   \includegraphics[width=0.85\linewidth]{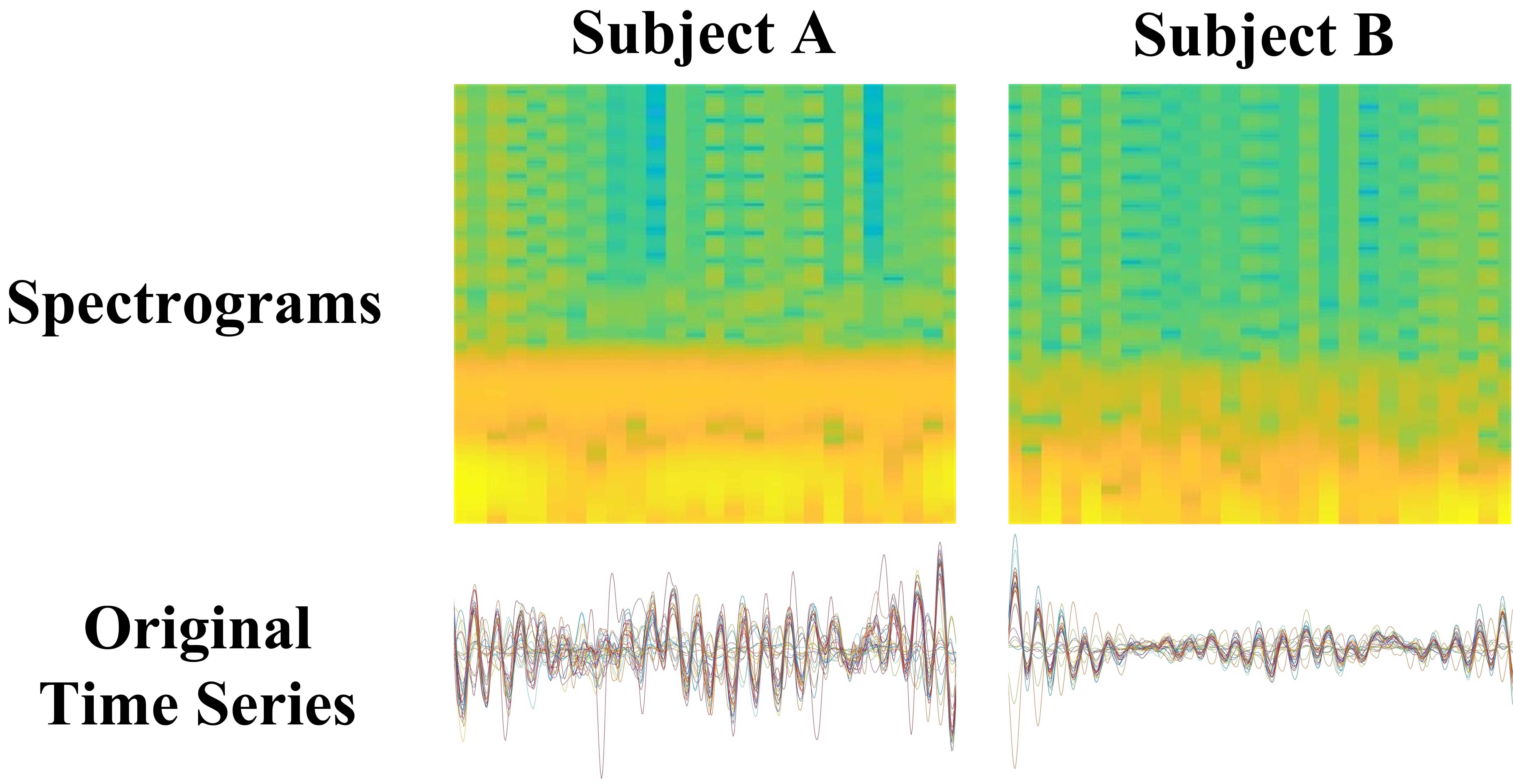}

   \caption{
   Time series and spectrograms of EEG signals from two subjects, both derived from the same visual stimulus.
   }
   \label{fig:onecol}
\end{figure}

Previous methods integrating Long Short-Term Memory (LSTM) models and Generative Adversarial Networks (GAN) have faced limitations due to the scarcity of training data for GANs, resulting in subpar semantic accuracy and image quality~\cite{218, 220, 221}. A novel approach employing Masked Autoencoders (MAE)~\cite{217} and latent diffusion model (LDM)~\cite{211} showed promise with improved reconstruction outcomes. However, this strategy encounters its own set of challenges: (1) poor feature embedding capability, relying on extra datasets for better representation; (2) neglecting frequency features as shown in Fig. ~\ref{fig:onecol}, which possess unique characteristics divergent from those of the time domain~\cite{supply-10}; (3) only coarse-grained category reconstruction, missing fine-grained semantic details.

To overcome these limitations, we propose \textbf{BrainVis}, a novel pipeline for EEG-based image reconstruction. It comprises three components: (1) EEG encoder, including self-supervised latent masked modeling and frequency features to enhance EEG representations; (2) Semantic interpolation cross-modal alignment, aligning EEG embeddings with coarse and fine-grained semantics in the CLIP~\cite{209} space to highlight primary visual components and reduce alignment difficulty; (3) Cascaded diffusion models, using aligned fine-grained EEG embeddings and coarse-grained classification results for semantic reconstruction and refinement.

\begin{figure*}[t]
  \centering
  \includegraphics[width=\textwidth]{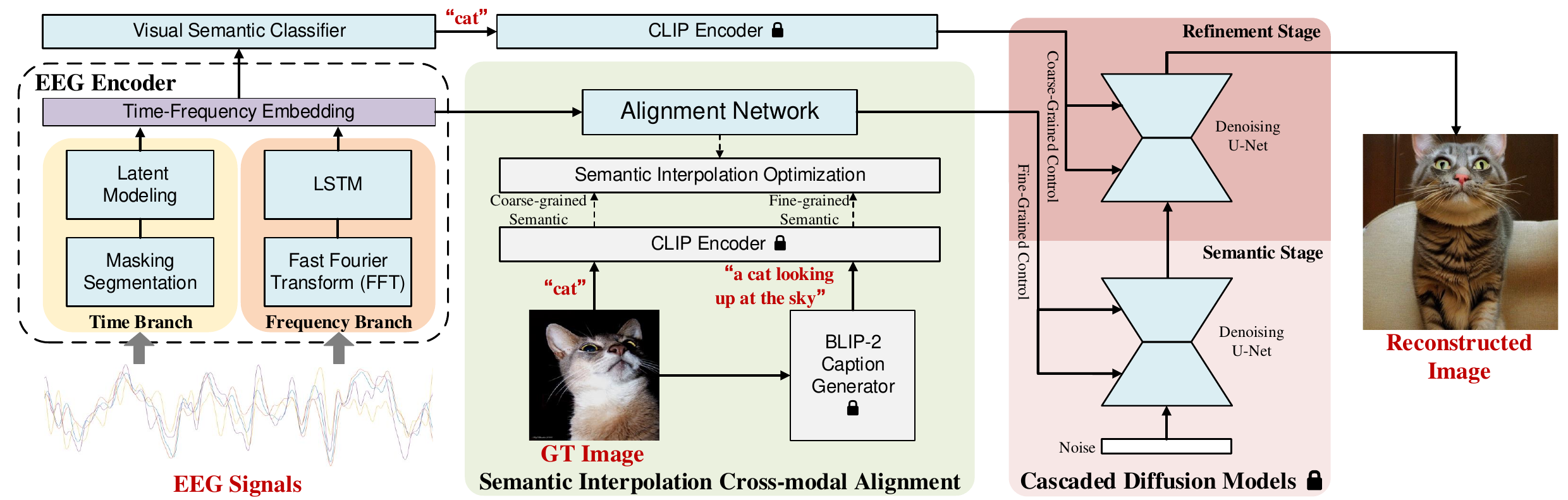}
  \caption{Framework of our proposed \textbf{BrainVis}. The blue blocks are the components including in inference process of the pipeline. The ground truth (GT) images are only used for supervision during training.}
  \label{fig:main}
\end{figure*}

\begin{figure}[t]
  \centering
   \includegraphics[width=\linewidth]{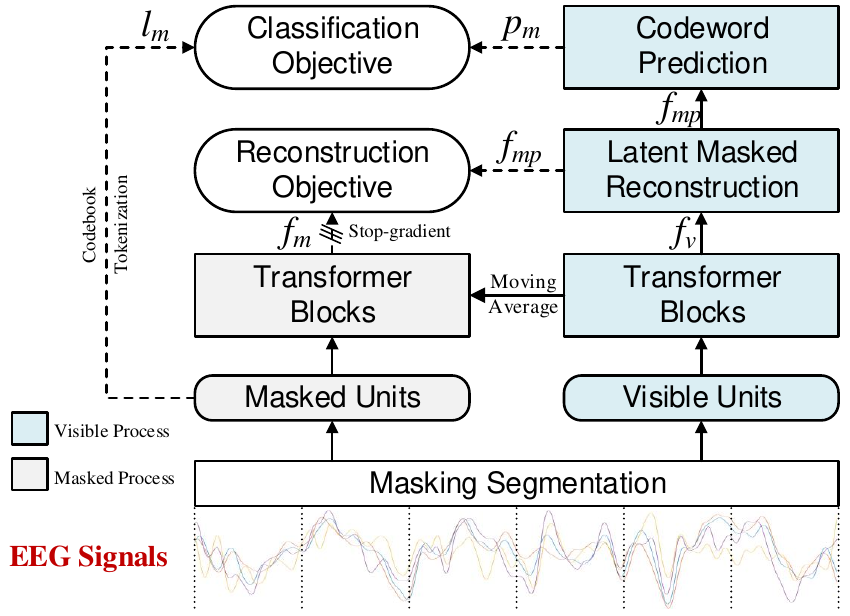}

   \caption{Latent Masked Modeling (LMM) of time branch. First, the EEG is segmented into visible and masked slices, then the masked slices are tokenized as $ l_m $. Next, the visible features $ f_v $ are extracted by transformer blocks, which are used for predicting the masked features $ f_{mp} $ and their codewords. The probability representation of predicted codewords is $ p_m $. Meanwhile, a non-trainable model obtained by moving average from the transformer blocks is used to extract real features of masked slices $ f_m $. Finally, reconstruction and classification objectives are performed.}
   \label{fig:time-enoder}
\end{figure}

Our contributions can be summarized as:
\begin{itemize}
    \item We improve the self-supervised embedding method for EEG time-domain features, and introduce the frequency features to enhance EEG representations. 

    \item We eliminate the reliance on additional large-scale datasets, reducing the data usage from 132k to 12k.

    \item BrainVis generates images with higher accuracy and better quality, and also overcomes previous limitation that was limited to only coarse-grained reconstruction.
\end{itemize}

The experimental results demonstrate that our proposed BrainVis outperforms the state-of-the-art (SOTA) in both semantic reconstruction and generation quality.

\section{Related Works}
\subsection{Reconstructing Visual Stimuli from fMRI}
Functional Magnetic Resonance Imaging (fMRI) boasts higher spatial resolution, better informative capacity and lower noise than EEG, facilitating its prevalence in visual reconstruction efforts~\cite{107}. Early studies~\cite{201,202,203} leveraged linear regression alongside traditional or DL features for visual categorization or reconstruction through deep decoding networks~\cite{204}. With the advent of GANs~\cite{205}, a shift towards employing GANs for visual reconstruction has emerged~\cite{206, 207}.

Recently, a study~\cite{208} used pre-trained CLIP model~\cite{209} to obtain embeddings of images and their artificial captions, aligning fMRI with these CLIP embeddings, then used it as the conditions for image generation by StyleGAN2. A study~\cite{210} transferred this latent embedding alignment to LDM~\cite{211}, achieving higher-quality reconstruction. Subsequently, works that centered around CLIP and LDM~\cite{212,213,214,215} continued to improve the accuracy and image quality of visual reconstructions, including introducing vector quantized-variational autoencoder (VQ-VAE)~\cite{216} and masked brain modeling (MBM) based on MAE~\cite{217} for better fMRI feature embedding. However, applying fMRI in practical settings is limited due to high cost, large data scale, and high processing complexity.

\subsection{Reconstructing Visual Stimuli from EEG}

EEG, by contrast, is more accessible and economically viable than fMRI. Nonetheless, its lower spatial precision and inherent noise considerably complicate the reconstruction of visual stimuli. Early studies~\cite{218} used LSTM for EEG classification and subsequently employed variational autoencoders (VAE) and GAN for image generation from EEG features, demonstrating the potential of EEG in conveying visual semantics. However, the feasibility of such attempts, constrained by EEG's intrinsic limitations and the small size of EEG-Image pair datasets, remained limited under frameworks utilizing GAN~\cite{220, 221}.

Inspired by MBM~\cite{213}, the recently proposed DreamDiffusion~\cite{222} embeds time-domain features of EEG using MAE and aligns them with CLIP-encoded images, as conditions of pre-trained LDM to achieve coarse-grained category reconstruction. However, it only captures the time-domain features of EEG signals, and the ability of its models' unsupervised feature embedding relies on pre-training on an additional large-scale self-collected EEG dataset (approximately 120,000 samples), indicating that it requires further enhancement in the representation capability of EEG features. Moreover, the condition of well pre-trained stable diffusion is text embedding, but DreamDiffusion directly aligns EEG features with image embeddings in CLIP space, which may introduce additional semantic noise, since text and image embeddings are not entirely equivalent.

\section{Method}
\label{sec:methods}


\subsection{Preliminary}
\noindent\textbf{CLIP}~\cite{209} is a pre-trained model for image-text similarity measurement, which embeds them in the same latent space with the contrastive learning. It mainly consists of an image encoder and a text encoder.

\noindent\textbf{Stable diffusion}~\cite{211} is a text-to-image generation model, which uses CLIP embedding of text as its conditional input. It guides the denoising process from random noise by controlling the conditional input of the U-Net, realizing the generation of images with specified semantics.

\noindent\textbf{BLIP-2}~\cite{302} is a visual-language pre-trained model that can generate captions from images. It utilizes pre-trained image and language models with fixed parameters, and uses a query transformer to fill the gap between their modalities.

\subsection{Overview}
In Figure \ref{fig:main}, we present BrainVis, which comprises three components: EEG encoder, semantic interpolation cross-modal alignment module, and cascaded diffusion models. The EEG encoder leverages self-supervised latent masked modeling to extract time-domain features and uses LSTM with Fast Fourier Transform (FFT) for frequency-domain features. The time and frequency branches are fine-tuned jointly with a visual semantic classifier to obtain time-frequency embeddings. These embeddings are then aligned with the semantic interpolation of CLIP embeddings, which represent visual stimuli labels and fine-grained captions generated by BLIP-2~\cite{302}. Finally, cascaded diffusion models utilize the aligned embeddings for semantic reconstruction and the classification results for semantic refinement.

\subsection{EEG Encoder}
\label{sec:time-fre}
\textbf{Latent Masked Modeling (LMM) Pre-training.} To learn EEG time-domain features, prior work~\cite{222} utilized MAE~\cite{217} for mask modeling, reconstructing the entire signals in the time domain directly. However, it demands a substantial volume of training data and struggles to represent complex EEG accurately. Drawing inspiration from the prior art~\cite{301}, we reconstructs the high-dimensional latent embeddings of single masked unit of the signal, increasing the feature density to reduce the learning difficulty and decrease data dependency.

Technically, as shown in the Latent Masked Modeling of Time Branch part of Figure \ref{fig:main}, the EEG signal $ x \in \mathbb{R}^{c \times l} $ is uniformly divided into $ n $ units. Each unit is projected into a $d$-dimension space to extract local features across channels, resulting in a representation $ z \in \mathbb{R}^{n \times d} $. A random masking strategy with a mask ratio $ r_m $ partitions $ z $ into visible and masked parts. The self-supervised learning objective has two goals to enhance feature representations: reconstructing unit embeddings and codebook-based unit classification.

For reconstruction objective, we first use transformer blocks to learn visible features $ f_v $. Masked features $ f_m $ are obtained by a non-trainable model derived from $ f_v $ transformer blocks via moving average. Further, new transformer blocks use $ f_v $ to predict $ f_m $, the predictive value is $ f_{mp} $. The mean squared error (MSE) defines the regression loss $ L_{reg} $:
\begin{equation}
  L_{reg}=\frac{1}{d}{||f_m-f_{mp}||_2^2}.
\end{equation}

For classification objective, a tokenizer assigns a one-hot codeword $ l_m $ to each masked unit as self-supervised labels via the codebook. The tokenizer consisting of linear layers, whose highest output determines the codebook index for the units. The total number of codewords is $ n_t $. By classifying $ f_{mp} $, we get the probability representation $ p_m $. The cross-entropy (CE) error defines the classification loss $ L_{cls} $:
\begin{equation}
  L_{cls}=-l_m\cdot \log(p_m)
\end{equation}
The overall learning objectives of the LMM can be defined as:
\begin{equation}
  L_{lmm}=L_{reg}+L_{cls}.
\end{equation}

\noindent\textbf{Time-Frequency Embedding (TFE).} In frequency branch, we convert EEG to frequency domain using Fast Fourier Transform (FFT). We observe that learning frequency features through complex networks risks overfitting. Therefore, we use an LSTM model in frequency branch. EEG visual labels are used as supervision to train the LSTM with CE loss. Then, time and frequency branches are fine-tuneing together with CE loss to build the unified time-frequency embedding.

\subsection{Semantic Interpolation Cross-modal Alignment}
To achieve high-quality image reconstruction, we use a pre-trained stable diffusion~\cite{211} model as the backbone generator. Previous work~\cite{222} aligned EEG embeddings with image CLIP embeddings before inputting them to the stable diffusion (SD) model. However, SD requires textual descriptions as input, and due to the gap between image and text CLIP embeddings \cite{zhou2022towards}, they failed to reconstruct fine-grained semantic. To adapt SD for our task, we align EEG signal embeddings with text semantic embeddings. 

For each instance, we only have coarse category annotations like “\emph{cat}” and its CLIP embeddings $ c_{label} $, which dose not capture the complex semantics in the image. To incorporate more fine-grained semantics, we generate pseudo captions using BLIP-2~\cite{302}. We then obtain the CLIP embeddings of the captions $ c_{cap} $ through CLIP encoder. However, aligning EEG features with $ c_{cap} $ is challenging. To address this, we align EEG features with the interpolation of fine and coarse-grained semantics, represented by generated captions and category annotations, respectively. The semantic interpolation aims to blend coarse and fine-grained semantics smoothly, and $ c_{label} $ highlights the primary semantic components of $ c_{cap} $ to ease the alignment difficulty.

We use a network composed of fully connected layers with residual connections to achieve this. Technically, we maximize the cosine similarity between the output of the alignment network $ c_{EEG} $ and $ c_{label} $ and $ c_{cap} $. The loss function $ L_{si} $ for semantic interpolation alignment is defined as:
\begin{equation}
  L_{si}=2-\frac{c_{cap}\cdot c_{EEG}}{|c_{cap}||c_{EEG}|}-\frac{c_{label}\cdot c_{EEG}}{|c_{label}||c_{EEG}|}
\end{equation}

\subsection{Cascaded Diffusion Models}

Due to the inevitable imperfect alignment, there exist information loss and semantic noise in our reconstruction process. To mitigate the negative impact, we propose cascaded diffusion models. We adopt different levels of semantics as conditional inputs for the pre-trained stable diffusion~\cite{211} model to guide the image reconstruction process. 

First, we use $ c_{EEG} $ as condition, performing reverse diffusion in the latent space to reconstruct the semantics images. However, the reconstructed images are of low quality due to semantic noise and information loss in $ c_{EEG} $. Therefore, we refine the incomplete denoised latents from reverse diffusion process by using unambiguous EEG classification labels as the new condition, which are from Sec.\ref{sec:time-fre}. It reduces the semantic noises and yields higher-quality results.


\section{Experiments}
\subsection{Dataset and implementation}
\noindent\textbf{Dataset.} Our experimental setup employs an EEG-image pairs dataset named Block-Design Visually-Evoked (BDVE)~\cite{223, supply-5,supply-6}, encompassing 11,466 EEG sequences recorded from 128-channel electrodes, filtered through a 5-95 Hz filter. These sequences were associated with 40 image classes. To elicit neural responses, six subjects were presented with a total of 2000 object images. These visual stimuli were selected from a subset of the widely used ImageNet~\cite{401} dataset and comprised 40 classes, each containing 50 easily recognizable images. To ensure the exclusion of potential interference from previously shown images, the first 40 ms of each recorded EEG sequence were discarded, and the following 440 ms of EEG data were utilized for the experiments. We follow the official~\cite{223, supply-5} and baselines~\cite{218,222}, split the dataset into training, validation, and testing sets using an 8:1:1 ratio based on images instead of subjects, avoiding semantic overlap.

\noindent\textbf{Implementation details.} We follow the optimal parameter settings of prior works~\cite{213, 222} and empirically set our parameters in Section \ref{sec:methods}: $ c=128 $, $ l=440 $, $ n=110 $, $ d=1024 $, $ r_M=0.75 $ and $ n_t=660 $. Moreover, the pre-trained stable diffusion model is v1-5-pruned-emaonly, while the pre-trained CLIP encoder is ViT-L/14. Transformer for visible feature extraction consists of 8 self-attention blocks, and for latent masked reconstruction comprises 4 cross-attention blocks. In each block, the attention heads are set to 16, and feed-forward network dimension is fixed at 4096. Additionally, the size of LSTM is 128. 
We employ a learning rate of 0.001 with a batch size of 128. Pre-training epochs are 300 for time encoder and 900 epochs for frequency encoder. Fine-tuning for time encoder lasts 80 epochs, and joint fine-tuning for time and frequency encoder lasts 30 epochs. Cross-modal EEG alignment fine-tuning is performed for 200 epochs.


\subsection{Evaluation metrics}
\textbf{EEG Classification Accuracy (CA).} CA measures the accuracy of EEG classification. It reflects the extent to which EEG's latent features are correlated with coarse-grained visual semantics.

\noindent\textbf{$N$-way Top-$K$ Classification Accuracy of Generation (GA).} Following~\cite{213, 222}, we utilize $N$-way Top-$K$ accuracy to evaluate the semantic correctness of the reconstruction results. For multiple trials, the top-$K$ and classification accuracies are calculated based on $ N-1 $ randomly selected classes, along with the correct class.
Both the ground-truth and generated images are initially inputted into a pre-trained ImageNet1K classifier~\cite{402}. We then examine whether the top-$K$ classification among the $ N $ selected classes matched the ground-truth classification. In our case, we set $ N=50 $ and $ K=1 $.

\noindent\textbf{Inception Score (IS).} IS is employed to evaluate the diversity of the reconstructed images, providing insights into their quality. For a detailed definition, please refer to~\cite{403}.

\noindent\textbf{Structural Similarity Index Measure (SSIM).} SSIM quantifies the similarity between images based on brightness, contrast, and structure. It is a low-level measure of image quality.

\noindent\textbf{F1-Score.} F1-Score is the harmonic mean of precision and recall, used to measure EEG classification performance. 

\begin{figure*}[t]
  \centering
  \includegraphics[width=\linewidth]{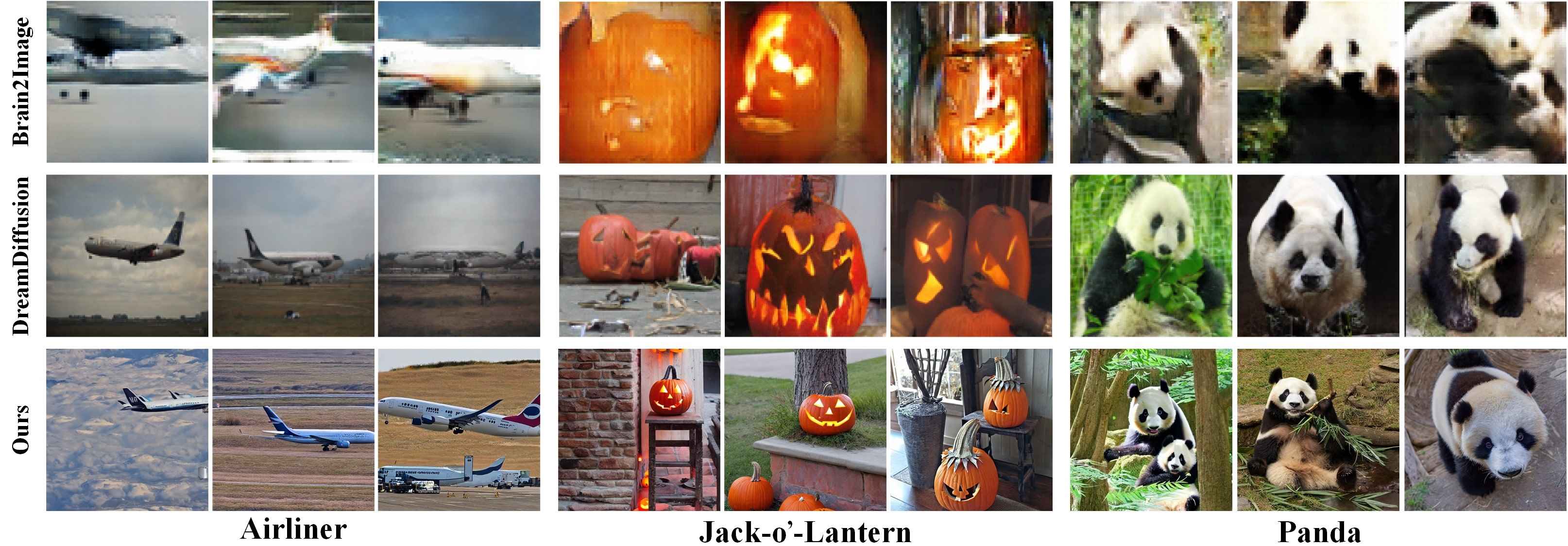}
  \caption{Comparison of reconstructed images’ quality.}
  \label{fig:comp1}
\end{figure*}

\begin{figure*}[t]
  \centering
  \includegraphics[width=\linewidth]{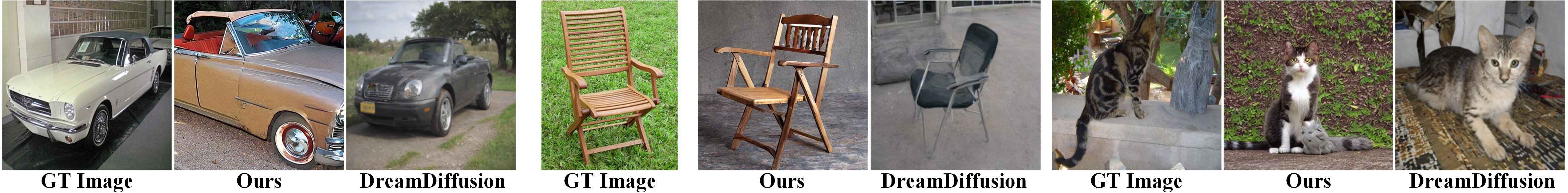}
  \caption{Comparison of reconstructed images with DreamDiffusion on the same ground truth (GT) image.}
  \label{fig:comp2}
\end{figure*}

\begin{table}
  \centering
  \footnotesize
  \begin{tabular}{l|c|ccc}
    \toprule
    Methods & Subject & GA $\uparrow$ & IS $\uparrow$ & FID $\downarrow$\\
    \midrule
    DreamDiffusion & \textbf{4} & 0.4580 & - & -\\
    \midrule
    BrainVis (Ours) & \textbf{4} & \textbf{0.4927}&31.5281&121.0212\\
    \midrule
    \midrule 
    Brain2Image &  & - & 5.0700 & -  \\
    NeuroVision&  & - & 5.1500 & -  \\
    DCLS-GAN& \textbf{Average} & - & 6.6400 & -  \\
    ESG-ADA&  & - & 10.8200 & 174.1300 \\
    MB2C&  & - & 12.8500 & 153.3700 \\
    \midrule
    BrainVis (Ours) & \textbf{Average} & 0.4546&\textbf{30.9985}&\textbf{126.6576}\\
    \bottomrule
  \end{tabular}
  \caption{Comparison with image reconstruction baselines. '-' are not publicly avaliable.}
  \label{tab:comp1}
\end{table}

\setlength{\tabcolsep}{3pt}
\begin{table}[t]
  \centering
  \footnotesize
  \setlength{\tabcolsep}{1pt}
  \begin{tabular}{l|cccccccccccc}
    \toprule
    Method & Top-1 CA $\uparrow$ & Top-3 CA $\uparrow$ & Top-5 CA $\uparrow$ & F1-Score $\uparrow$  \\
    \midrule
    EEG-CN & 0.3130 & - & - & - \\
    Brain2Image&0.1648 &0.3790 &0.5524 & 0.1607\\
    S-EEGNet  & 0.2520 & - & - & - \\
    KD-STFT & 0.4120 & 0.7530 & 0.8782 & 0.4027\\
    \midrule
    BrainVis (Ours) & \textbf{0.4766} & \textbf{0.7923} & \textbf{0.9120} & \textbf{0.4323}\\
    \bottomrule
  \end{tabular}
  \caption{Comparison with EEG classification baselines. '-' are not publicly avaliable.}
  \label{tab:comp2}
\end{table}

\subsection{Comparison against other methods}
\label{sec:results}
Our comparative experiments include image reconstruction and EEG classification. The baselines of image reconstructions include DreamDiffusion~\cite{222}, Brain2Image~\cite{218}, NeuroVision~\cite{nv}, DCLS-GAN~\cite{dcls}, ESG-ADA~\cite{ada} and MB2C~\cite{mb2c}. Notably, DreamDiffusion only reports GA on subject 4. The baselines of EEG classification include EEG-CN~\cite{supply-5}, Brain2Image~\cite{218}, S-EEGNet~\cite{supply-8} and KD-STFT~\cite{supply-7}. Furthermore, since we aim to reconstruct semantics, low-level metrics are not considered.

\noindent\textbf{Quantitative results.} 
To compare image reconstruction capabilities, we conduct experiments generating and evaluating 4 images per EEG sample for all subjects. For EEG representation capability, we conduct EEG classification experiments.

The results in Table \ref{tab:comp1} show that BrainVis surpasses the existing SOTA by 7.04\%, 186.49\%, and 27.26\% in GA, IS, and FID, respectively. It demonstrates that BrainVis has the best reconstruction semantic fidelity and generation quality.

The results in Table \ref{tab:comp2} show that BrainVis surpasses the existing SOTA by 15.68\%, 5.22\%, 3.85\%, and 7.35\% in Top-1, Top-3, Top-5 CA, and F1-Score, respectively. It demonstrates that our proposed EEG encoding (LMM+TFE) method has the best EEG representation capability.

\begin{figure*}[t]
  \centering
  \includegraphics[width=\linewidth]{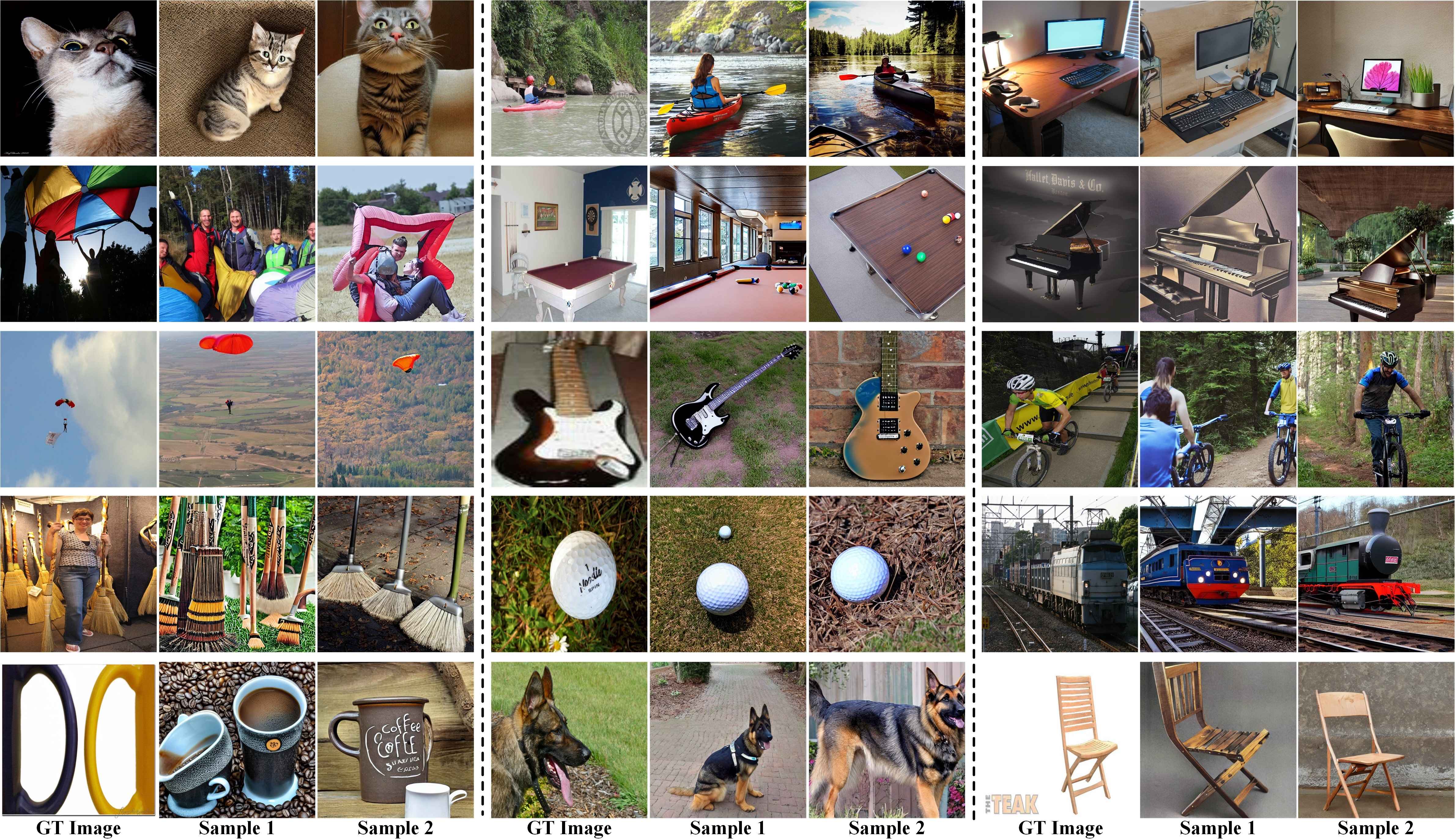}
  \caption{Image reconstruction results and the GT images. We not only reconstruct the correct types, but also capture the fine-grained semantics to some extent, such as prominent numerical, color, or behavioral features.}
  \label{fig:main_results}
    \vspace{-0.1 in}
\end{figure*}

\noindent\textbf{Qualitative results.} 
In Figure \ref{fig:comp1}, we compare our reconstructed images for three categories (Airliner, Jack-o’-Lantern, and Panda) with classic GAN-based Brain2Image and diffusion-based DreamDiffusion, which is widely considered as the SOTA in image reconstruction quality. Our images show higher quality, highlighting our method's effectiveness.

In Figure \ref{fig:comp2}, we compare our reconstructed images with DreamDiffusion using the same ground truth (GT) images. Due to the limited GTs publicly available from DreamDiffusion, only 3 images are available for comparison under consistent conditions. Among them, we correctly reconstruct the styles and colors of the car and chair, as well as the behavior and environment of the cat, while DreamDiffusion failed.

\begin{table}
  \centering
  \footnotesize
    \begin{tabular}{l|c|cccc}
    \toprule
    Models & Full & w/o TiB & w/o FrB & w/o PT & w/o FT\\
    \midrule
     Top-1 CA $\uparrow$ & \textbf{0.4766}&0.2943&0.4397&0.2783&0.2495\\
    \bottomrule
  \end{tabular}
  \caption{Ablation analysis of components of EEG encoder. TiB, FrB, PT and FT respectively stand for time branch, frequency branch, LMM pre-training and TFE fine-tuning.}
  \label{tab:ab2}
    \vspace{-0.1 in}
\end{table}

\begin{table}
  \centering
  \footnotesize
  \begin{tabular}{l|c|ccc}
    \toprule
    \multicolumn{2}{c|}{Models} & GA$\uparrow$ & IS$\uparrow$ & FID$\downarrow$\\
    \midrule
      &w/o Refinement&0.1218&16.5092&255.3962\\
     w/o FrB & w/o Semantic&0.4147&28.1505&137.8096\\
     &Cascaded&0.4202&29.5453&132.8096\\
     \midrule
      &w/o Refinement&0.1291&17.4441&244.2778\\
     w/o TiB &w/o Semantic&0.2782&28.5453&129.0044\\
     &Cascaded&0.2876&30.3657&127.6774\\
     \midrule
     &w/o Refinement & 0.1307&17.8735&242.9537\\
     Full &w/o Semantic& 0.4501&28.8167&129.1161\\
     &Cascaded& \textbf{0.4546}&\textbf{30.9985}&\textbf{126.6576}\\
    \bottomrule
  \end{tabular}
    \caption{Ablation study on the time branch, frequency branch and cascaded models on image reconstruction.}
  \label{tab:ab3}
  \vspace{-0.1 in}
\end{table}

\begin{table}
  \centering
  \footnotesize
  \begin{tabular}{l|cccc}
    \toprule
    Pre-trained with & CA$\uparrow$& GA$\uparrow$ & IS$\uparrow$ & FID$\downarrow$ \\
    \midrule
    BDVE (Ours, 12k)&\textbf{0.4766}& \textbf{0.4546} & \textbf{30.9985}& \textbf{126.6576} \\
    MOABB (120k)&0.3413& 0.3127 & 29.0601 & 147.1737 \\
    BDVE+MOABB&0.4739& 0.4523 & 30.5221 & 126.7201 \\
    None&\textbf0.2783& 0.2386& 29.1481 &145.7066\\
    \bottomrule
  \end{tabular}
  \caption{Ablation study on pre-training data.}
  \label{tab:data}
\end{table}

\begin{table}
  \centering
  \footnotesize
  \begin{tabular}{l|ccccc}
    \toprule
    $n_t$ & None & 220 & 330 & \textbf{660} & 990\\
    \midrule
     Top-1 CA $\uparrow$ & 0.4118&0.4382&0.4326&\textbf{0.4397}&0.3597  \\
    \bottomrule
  \end{tabular}
  \caption{Ablation study on codewords number $n_t$ on CA (w/o the frequency branch).}
  \label{tab:ab1}
    \vspace{-0.1 in}
\end{table}

\begin{table}
  \centering
   \footnotesize
  \begin{tabular}{l|ccc}
    \toprule
     Model  & Time (s/epoch) $\downarrow$ & Params $\downarrow$ & Top-1 CA $\uparrow$ \\
    \midrule
     CNN &75.7974&5.1410M & 0.2537 \\
     LSTM&\textbf{13.5687}&\textbf{0.1538M} &\textbf{0.2578}\\
    \bottomrule
  \end{tabular}
  \caption{Comparison of training time, parameter numbers and CA between different frequency encoder backbones.}
  \label{tab:mot}
\end{table}

\subsection{Ablation Study}
\label{sec:as}
\noindent\textbf{Components of EEG encoder.}
In Table \ref{tab:ab2}, we conducted an ablation study through classification experiments to analyze the impact of the main components in the EEG encoder on EEG representation. The results show that time-domain features, frequency-domain features, LMM pre-training and time-frequency embedding positively affect performance.

\noindent\textbf{The impact of time branch, frequency branch
and cascaded models on image reconstruction.} The results in Table \ref{tab:ab3} show that time-domain features dominate image reconstruction, while frequency-domain features improve it. Additionally, each stage of the cascaded diffusion models positively contributes to the reconstruction process.

\noindent\textbf{Pre-training data.} Previous work used various datasets from MOABB~\cite{moabb}, including many EEG recordings unrelated to visual tasks. The number of EEG channels and their meaning vary, and EEGs with fewer than 128 channels are unscientifically filled by self-replication. Table~\ref{tab:data} shows that including these data in our pre-training may even slightly negatively impact performance.

\noindent\textbf{Optimization of codewords number.} In Table \ref{tab:ab1}, we present the results of ablation analysis on the total number of codewords $n_t$. Based on the results, we set $ n_t=660 $. 

\noindent\textbf{Backbone of frequency branch.} We observe that using complex networks in the frequency branch precipitates rapid overfitting. To mitigate this, we favor simpler backbones for the frequency branch, such as convolutional neural network (CNN) and LSTM, which are commonly used to process EEG. We choose the CNN proposed in~\cite{223} as an example. The results in Table \ref{tab:mot} show that using LSTM effectively reduces training costs and slightly improves performance.

\subsection{Image Reconstruction Results}
In Figure \ref{fig:main_results}, we present part of the reconstructed results. We can reconstruct the correct coarse-grained categories and preserve fine-grained details. For example, considering the first column, we successfully reconstruct a cat looking up, parachutes on the ground or in the air, and the quantitative relationship between brooms/cups and their GT images, etc. 

\subsection{Discussion}
\begin{figure}[t]
  \centering
   \includegraphics[width=\linewidth]{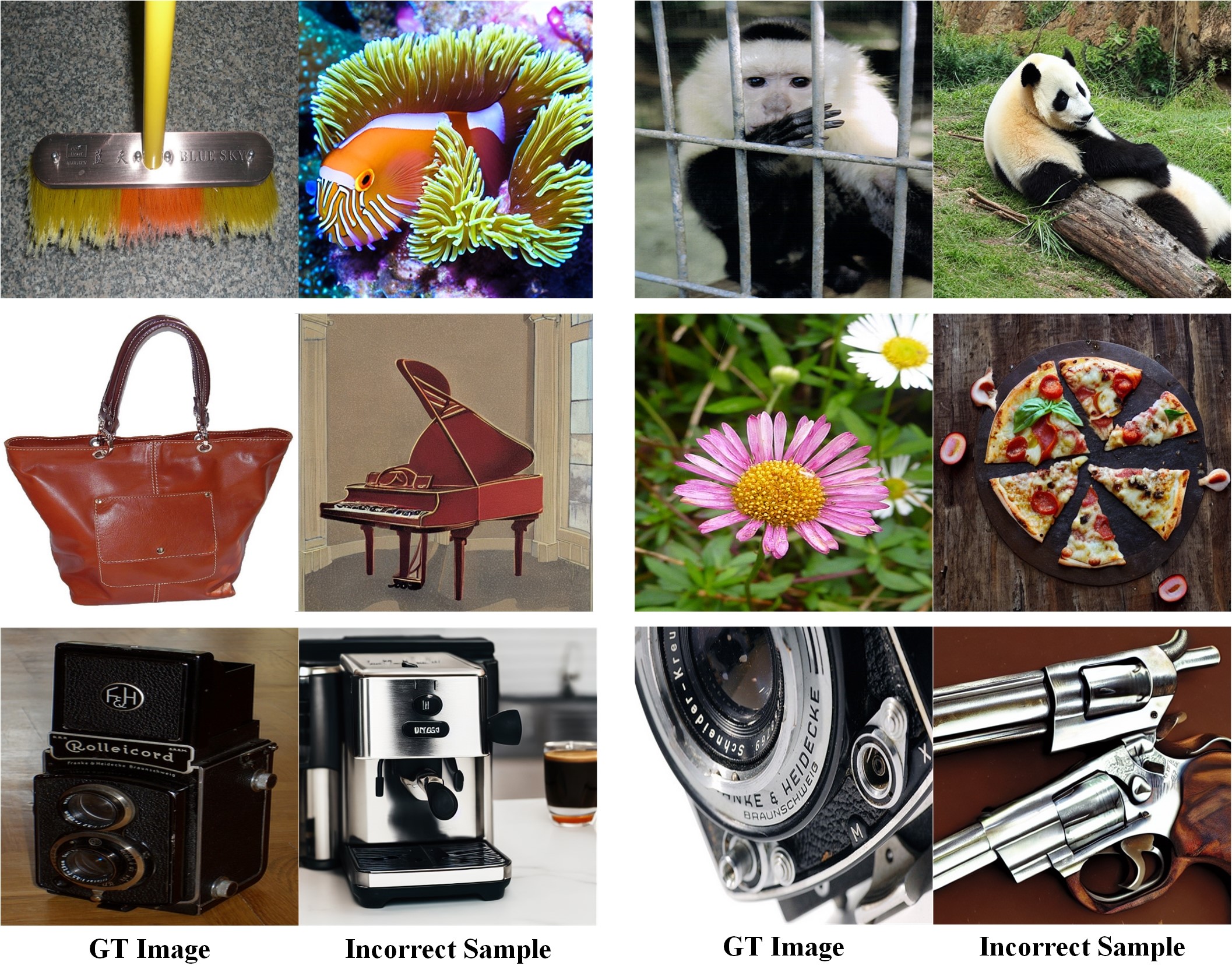}
   \caption{Examples of incorrect reconstruction results.}
   \label{fig:failure}
\end{figure}

In figure \ref{fig:failure}, we present part of the reconstruction results under incorrect coarse-grained classification. Although the generated object is incorrect, there are noticeable resemblances between these generated samples and real images, particularly in terms of color, shape or surface textures/materials. This observation implies that the visual information encoded in human EEG signals may prioritize fundamental properties of objects rather than specific object categories. The classification of visual stimuli from EEG signals involves decoding and categorizing information pertaining to color, shape and various other features. Building upon this hypothesis, decoding individual properties of visual information, such as color and shape in isolation may offer an effective avenue for EEG analysis and visual reconstruction.

\section{Conclusion}
This paper proposes BrainVis, a novel pipeline for EEG-based image reconstruction. BrainVis uses self-supervised latent masked modeling and LSTM to embed EEG as time-frequency features, which are then aligned with semantically interpolated CLIP embeddings. The aligned results and predicted category embeddings serve as conditions for cascaded diffusion models for image reconstruction. BrainVis surpasses the SOTA in EEG representation, semantic reconstruction and generation quality, eliminates the need for additional unrelated data, and overcomes the previous limitation of being unable to reconstruct fine-grained semantics.

\bibliography{aaai25}

\end{document}